\documentclass{esannV2}
\usepackage{graphicx}
\usepackage[linesnumbered, ruled, vlined]{algorithm2e}
\usepackage{booktabs}

\DeclareGraphicsExtensions{.eps}  
\graphicspath{{media/}} 

\usepackage[utf8]{inputenc}
\usepackage{amssymb,amsmath,array}

\begin{document}
\title{Graph Diffusion Counterfactual Explanation}

\author{David Bechtoldt and Sidney Bender
\thanks{This work was supported by BASLEARN-TU Berlin/BASF Joint Laboratory, co-financed
by TU Berlin and BASF SE.}
\vspace{.3cm}\\
%
Machine Learning Group, TU Berlin, 10587 Berlin, Germany
}

\maketitle

\begin{abstract}
Machine learning models that operate on graph-structured data, such as molecular graphs or social networks, often make accurate predictions but offer little insight into \textit{why} certain predictions are made. Counterfactual explanations address this challenge by seeking the closest alternative scenario where the model's prediction would change. Although counterfactual explanations are extensively studied in tabular data and computer vision, the graph domain remains comparatively underexplored. Constructing graph counterfactuals is intrinsically difficult because graphs are discrete and non-euclidean objects. We introduce Graph Diffusion Counterfactual Explanation, a novel framework for generating counterfactual explanations on graph data, combining discrete diffusion models and classifier-free guidance. We empirically demonstrate that our method reliably generates in-distribution as well as minimally structurally different counterfactuals for both discrete classification targets and continuous properties.
\end{abstract}

\section{Introduction}

Explainable AI (XAI) aims to make the behavior of complex machine-learning models transparent and interpretable for humans \cite{Samek}. Among existing approaches, counterfactual explanations have emerged as a particularly intuitive and actionable concept. They ask what minimal change to the input would lead to a different prediction, thereby providing insight into model behavior, robustness, and bias \cite{Counterfactuals}. 
Although early counterfactual explanation methods largely focused on tabular data, the computer vision community has pioneered generative approaches that better align with our goals of realism and interpretability. These vision methods generate counterfactuals in the latent space of a model instead of directly perturbing input features, yielding semantic modifications that remain in the data manifold \cite{TIME,Diffeomorphic}. For example, Jeanneret et al. (2024) \cite{TIME} propose \textit{TIME}: a method using a diffusion model with learned textual embeddings and inversion to generate counterfactuals that remain in-distribution and without access to model gradients. Conceptually, this echoes the idea by Dombrowski et al. (2024) \cite{Diffeomorphic}, which advocates optimization in a generator-induced latent coordinate system so that counterfactual trajectories remain on the data manifold. Transferring these ideas to graphs is non-trivial because graphs are discrete, combinatorial, and non-Euclidean. Gradients in input space are ill-defined, and naive edit search becomes intractable for larger structures such as molecules \cite{Graph_Counterfactuals, FreeGress}. Discrete diffusion models such as \textit{DiGress} \cite{DiGress} partly address this by defining Markov transitions over node and edge categories and learning to denoise them with a graph transformer. Beyond counterfactual reasoning, there also exist successful XAI methods for graphs, e.g., attribution-based approaches such as \textit{GNN-LRP} \cite{LRPGraph}, but here we focus exclusively on counterfactual explanations. 
Diffusion for graphs addresses part of this problem.

\begin{figure}[htbp]
  \centering
  \includegraphics[width=0.95\textwidth]{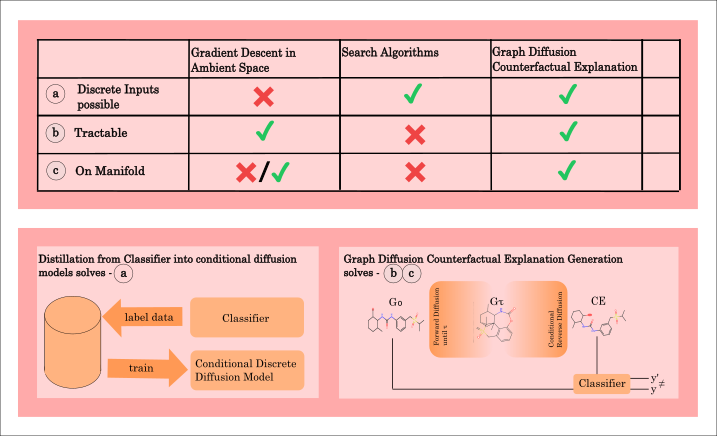}
  \caption{Top: Comparison across (a) support for discrete inputs, (b) tractable generation, and (c) on-manifold solution. Ambient-space gradient methods are computationally efficient, and some methods ensure data manifold closeness, like \textit{Diffeomorphic Counterfactuals} \cite{Diffeomorphic}, but assume differentiable, continuous inputs and therefore do not apply to categorical graph structure. Search-based methods respect discreteness but become combinatorial intractable on large graphs and do not enforce manifold constraints. Our method satisfies all three. Bottom-left: distillation of label information into a conditional discrete diffusion model. Bottom-right: \textit{GDCE} generation pipeline.}
  \label{fig:example-pipeline}
\end{figure}

We build on these insights and present a diffusion-based framework for graph counterfactuals that unifies on-manifold generation with tractable control (Fig.~\ref{fig:example-pipeline}). Concretely, we distill the dataset into a discrete graph diffusion model to circumvent the lack of gradients in discrete graph space, and we replace auxiliary-classifier guidance with classifier-free guidance for graphs introduced by Ninniri et al. (2024) \cite{FreeGress} by injecting the target condition directly into the generative model. During inference, we perturb the input graph partway along a diffusion trajectory and then guide reverse denoising toward the desired property, yielding counterfactual graphs that flip the prediction, remain close to the original graph, and respect the data manifold. Given the absence of directly comparable methods at large-scale datasets like \textit{QM9} and \textit{ZINC-250k}, we focus on within-method ablations and diagnostic checks. 

\section{Method: Graph Diffusion Counterfactual Explanation}

\textbf{Preliminaries}:
Let a graph \textit{G} = (\textit{X}, \textit{E}) be represented by node one-hots $\textbf{X} \in \mathbb{R}^{ n\times a}$ and edge type tensors $\textbf{E} \in \mathbb{R}^{ n\times n \times b}$. Discrete diffusion defines Markov transitions over nodes and edges via transition matrices $[Q^t_X]_{ij} = q(x^t=j|x^{t-1} = i)$ and $[Q^t_E]_{ij} = q(e^t=j|e^{t-1} = i)$ \cite{DiGress}. The forward process applies cumulative transitions $\bar{Q^t}=Q^1Q^2...Q^t$, producing noisy $\textit{G}_t = (\textit{X}\bar{Q^t}_\textit{X}, \textit{E}\bar{Q^t}_\textit{E})$. A neural denoiser parameterizes reverse transitions to sample $\textit{G}_{t-1}$ from $\textit{G}_t$.

\textbf{Classifier-free guidance for graphs}: We condition the denoiser on a target \textit{y} (discrete class or continuous property) with conditioning dropout training so the same network learns both conditional and unconditional denoising. At inference, we form a guided score by linearly combining conditional and unconditional predictions with scale \textit{s}. 
\textbf{Graph Diffusion Counterfactual Explanation}: Given an observed graph \textit{G} and a target $\textit{y}_1$, we forward perturb a sample $\textit{G}_\tau$ from $q(\textit{G}_\tau|\textit{G})$ for an intermediate step $\tau$. This erases fine details while preserving global structure. Secondly, we guide the reverse diffusion from \textit{t} = $\tau$ to 0 under the condition $\textit{y}_1$ using classifier-free guidance. The resulting Graph $\textit{G}_{CF}$ is our counterfactual. This design yields manifold-aware edits, meaning large-scale structure is retained by starting from $\textit{G}_\tau$, while the guided denoising injects just enough flexibility to reach the target with minimal edits. The algorithm can be contemplated in Algorithm~\ref{alg:method}.

\begin{algorithm}[ht]
\SetAlgoLined
\KwIn{A graph $G = (X, E, y_{0})$, with original feature $y_{0}$, target $y_{1}$, diffusion steps $\tau$}
\KwOut{A graph $G_{CE} = (X, E, y_{1})$}

Compute noisy graph $G_{\tau} \sim (X \bar{Q}_{X}^{\tau}, E \bar{Q}_{E}^{\tau})$\ 
\For{$t = \tau$ \KwTo $1$}{
    $\hat{p}_X, \hat{p}_E \leftarrow \phi_\theta(G_t)$ \tcp*[r]{Reverse pass}
    
    $p_\theta(x^{t-1}_i \mid G_t, y_{1}) \leftarrow \sum_x \, q(x^{t-1}_i \mid x^t_i, x^0_i = x) \cdot f_\theta(x^0_i = x \mid G_t, y_1)$ \;
    
    $p_\theta(e^{t-1}_{ij} \mid G_t, y_{1}) \leftarrow \sum_e \, q(e^{t-1}_{ij} \mid e^t_{ij},  e^0_{ij} = e) \cdot f_\theta(e^0_{ij} = e \mid G_t, y_1)$ \;

    $G_{t-1} \sim \prod_i p_\theta(x^{t-1}_i \mid G_t, y_1) \cdot \prod_{i,j} p_\theta(e^{t-1}_{ij} \mid G_t, y_1)$ \;
}
\KwRet \textit{$G = (X, E, y_{1})$}

\caption{Graph Diffusion Counterfactual Generation}
\label{alg:method}
\end{algorithm}

\section{Proof of Concept: Planar Graphs}

\textbf{Data:} We evaluate on all non-isomorphic planar graphs with 8 nodes. A graph is planar if it can be drawn without edge crossings on a plane. This domain lets us verify that outputs remain planar and quantify minimal structural change precisely. 

\textbf{Experimental setup}: 
We train a classifier-free guided discrete diffusion model conditioned on the edge count $y$. For each test graph $G$, we request a counterfactual with $y_{\text{target}} = |E| - 1$, generating graphs with exactly 8 nodes and the prescribed edge count. Per input, we sample 100 counterfactuals across $\tau \in \{1, 5, 10, 25, 50, 100, 200\}$. We report (i) structural validity as connected and planar, (ii) target accuracy among valid graphs, and (iii) mean Graph Edit Distance to the original. As a baseline, \textit{FreeGress} generates 100 samples under the same $y_{\text{target}}$.

\begin{table}[ht]
\centering
\caption{Graph Diffusion Counterfactual Generation for Planar Graphs}
\label{tab:results}
\begin{tabular}{lccccccccccc}
\toprule
\textbf{$y_{\text{target}} = |E| - 1$} & \textbf{FreeGress} & \textbf{$\tau$=1} & \textbf{$\tau$=5} & \textbf{$\tau$=10} & \textbf{$\tau$=50} & \textbf{$\tau$=100} & \textbf{$\tau$=200} \\
\midrule
Validity & 0.78 & 0.14 & 0.55 & 0.87 & \textbf{0.91} & 0.89 & 0.79\\
Accuracy & \textbf{1.00} & 0.05 & 0.67 & 0.96 & \textbf{1.00} & \textbf{1.00} & \textbf{1.00}  \\
Mean-GED & 4.13 & 1.00 & 2.54 & 2.43 & \textbf{1.57} & 2.13 & 3.95 \\
\bottomrule
\label{tab:results_planar}
\end{tabular}
\end{table}

\textbf{Results \& discusion:} Table \ref{tab:results_planar} 
shows that our method yields consistently high structural validity, especially for $\tau \ge 50$ and generally outperforms \textit{FreeGress}. Target accuracy is near-perfect at $\tau \ge 50$, while mean GED remains small (1.57).
Performance decreases beyond certain noise levels, reflecting the trade-off between edit capacity and closeness. Overall, the approach produces structurally valid counterfactuals with high accuracy and minimal deviations from the original.

\section{\textit{ZINC-250k}: Steering Molecules into Drug-like logP Ranges}

\textbf{Data:} 
We use \textit{ZINC-250k}, a dataset of 250,00 drug-like molecules with up to 38 heavy atoms and a broad set of physicochemical properties.
We target logP, the octanol-water partition coefficient, which governs the hydrophobicity and is central to the drug-likeness. We encode membership in a desirable interval $[1.7, 3.8]$, following the QED desirability range \cite{QED}.

\textbf{Experimental setup}: 
We train a classifier-free guided diffusion model conditioned on a binary indicator of whether a molecule's logP lies in $[1.7, 3.8]$. For evaluation, we sample 200 test molecules outside the target range covering both overly hydrophilic and hydrophobic cases and generate 10 counterfactuals per molecule at perturbation levels $\tau = 10, 50, 100$. We compare against \textit{FreeGress}.  

We evaluate the outcomes using (i) structural validity, the percentage of generated molecules that are chemically valid as determined by RDKit santization checks ensuring no valence or atom-type violations, (ii) target accuracy, the percentage of valid outputs that successfully have their logP within the desired [1.7, 3.8] range and (iii) structural similarity, a measure of how close the counterfactual molecules are to the original structure. For the smaller planar graphs, we used mean GED to quantify minimal changes. However, exact GED is computationally intractable for larger graphs since it is NP-hard. Instead, for \textit{ZINC-250k} we report the mean Tanimoto similarity between the original and generated molecules based on their Morgan fingerprint representations. 

\begin{table}[ht]
\centering
\caption{Graph Diffusion Counterfactual Generation for \textit{ZINC-250k} Molecules targeting desirable logP ranges}
\label{tab:results}
\begin{tabular}{lccccccccccc}
\toprule
\textbf{$logP \notin [1.7,3.8] \;\rightarrow\; logP \in [1.7,3.8]$} & \textbf{FreeGress} & \textbf{$\tau$=10} & \textbf{$\tau$=50} & \textbf{$\tau$=100}\\
\midrule
Validity & \textbf{0.71} & 0.59 & 0.40 & 0.16 \\
Accuracy & \textbf{0.55}& 0.32 & 0.41 & 0.50 \\
Similarity & 0.04 & \textbf{0.50} & 0.44 & 0.28 \\
\bottomrule
\label{tab:results_zinc_logP}
\end{tabular}
\end{table}

\begin{figure}[ht]
  \centering
  \includegraphics[width=0.95\textwidth]{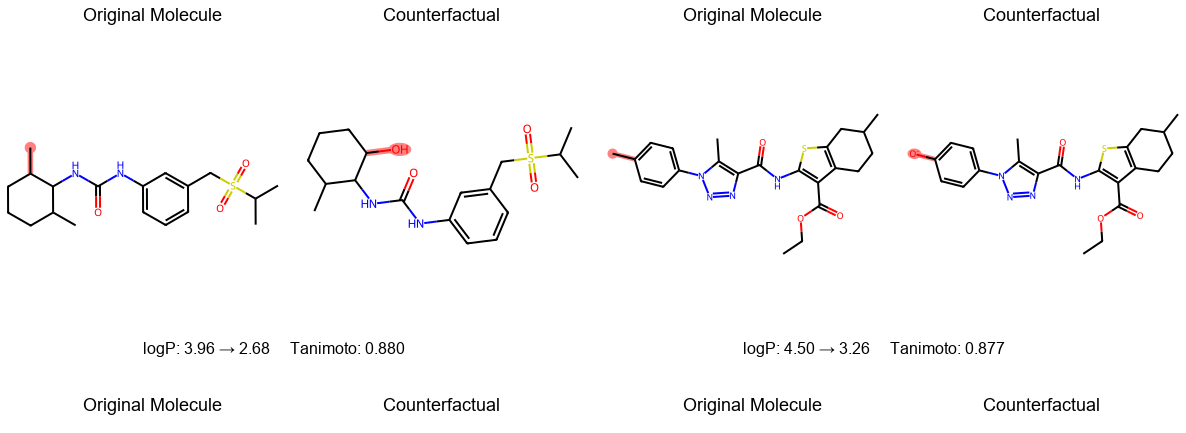}
  \caption{Two \textit{ZINC-250k} counterfactual examples that steer a molecule's logP into the prespecified target range.}
  \label{fig:zinc-examples}
\end{figure}

\textbf{Results \& discussion}: 
Table \ref{tab:results_zinc_logP} summarizes the performance of our guided diffusion approach versus \textit{FreeGress} in steering \textit{ZINC-250k} molecules into the desired logP range, and Figure~\ref{fig:zinc-examples} showcases \textit{ZINC-250k} counterfactuals. A clear trade-off emerges between achieving the target property and preserving the original structure. At low noise perturbation $\tau = 10$, our method produces counterfactuals with high structural similarity to the original molecule, with a mean Tanimoto similarity of 0.5, demonstrating that only minimal edits were made, but in comparison to the baseline with hampered validity of 59\% vs. 71\% and accuracy of 32\% vs. 55\%. As we allow larger perturbations, our approach becomes more effective at altering logP with 50\% at $\tau = 100$, but this comes at the cost of drastically reduced validity of only 16\% and reduced Tanimoto similarity of 0.28. These trends are intuitively consistent with the task difficulty that larger structural changes are more likely to achieve a challenging property target, yet they risk breaking chemical rules or straying too far from the original molecule. In contrast, the \textit{FreeGress} baseline, which freely generates new structures, to satisfy the property, achieves higher validity and accuracy overall but produces molecules essentially unrelated to the input, as expected. 

\section{Conclusion}

We presented Graph Diffusion Counterfactual Explanation, a classifier-free guided discrete diffusion framework for generating counterfactuals on graphs. The method integrates partial-noising edits with guided denoising, yielding on-manifold modifications that respect domain constraints while steering predictions toward desired targets. Across synthetic planar graphs and large drug like molecules from \textit{ZINC-250k}. Our method produced valid, accurate, and structurally close alternatives. As expected, results reveal a fundamental trade-off: tightening similarity constraints limits achievable property shifts, whereas allowing larger perturbations improves target success at the cost of similarity and, if pushed too far, validity. Crucially, the diffusion prior helps navigate this tension by biasing edits toward realistic, high-probability regions of the graph manifold. Methodologically, our approach unifies discrete changes and continuous property conditioning within a single, streamlined generator. 
Overall, these findings establish diffusion-based counterfactuals as a practical and scalable tool for interrogating and shaping model behavior in the graph domain, enabling users to ask \textit{what needs to change} and obtain actionable, domain-valid edits.

\begin{footnotesize}




\bibliographystyle{unsrt}
\bibliography{references.bib}

\end{footnotesize}


\end{document}